\documentclass[12pt,a4paper]{article}
\usepackage{hyperref}
\usepackage{amsfonts}
\usepackage{amsmath}
\usepackage{algorithm}
\usepackage{algpseudocode}
\usepackage{breakcites}
\usepackage[square]{natbib}   %
\newif\ifpdf
\ifx\pdfoutput\undefined
\pdffalse %
\else
\pdfoutput=1 %
\pdftrue
\fi

\ifpdf
\usepackage[pdftex]{graphicx}
\else
\usepackage{graphicx}
\fi

\ifpdf
\DeclareGraphicsExtensions{.pdf, .jpg}
\else
\DeclareGraphicsExtensions{.eps, .jpg}
\fi

\begin{document}

\thispagestyle{empty}
{\vspace*{-2.8cm} \hspace*{7.5cm} \includegraphics[width=75mm,viewport=0 0 235 85,clip]{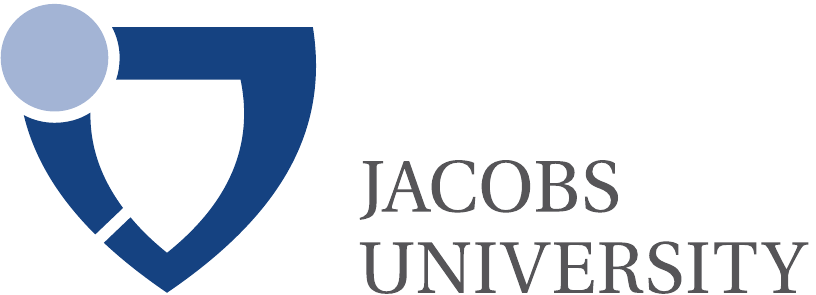}}
\vspace{6.5cm}

{\noindent \large \textsf{Xu He, Herbert Jaeger}} \\
\\

{\noindent\LARGE \bf \textsf{Overcoming Catastrophic Interference by Conceptors}} \\
\vfill

{\noindent \LARGE \textsf{Technical Report No. 35}} \\
{\noindent \textsf{May 2017}} \\
{\hspace*{-3.17cm} \rule[3mm]{\textwidth}{0.75pt}} \\
{\LARGE \textsf{School of Engineering and Science}} \\

\newpage
\thispagestyle{empty}

{ 
\noindent
\huge {\bf Overcoming Catastrophic Interference by Conceptors} \\
\\
\normalsize

\vspace{1cm}

\noindent
{\bf Xu He, Herbert Jaeger}\\
\\
{\it 
School of Engineering and Science\\
Jacobs University Bremen gGmbH\\
Campus Ring 12\\
28759 Bremen\\
Germany\\ 
\\
E-Mail: \href{mailto:x.he@jacobs-university.de}{x.he@jacobs-university.de}\\
\url{http://minds.jacobs-university.de/} 
} \\

}

\section*{Abstract}
Catastrophic interference has been a major roadblock in the research of continual learning. Here we propose a variant of the back-propagation algorithm, ``conceptor-aided back-prop" (CAB), in which gradients are shielded by conceptors against degradation of previously learned tasks. Conceptors have their origin in reservoir computing, where they have been previously shown to overcome catastrophic forgetting. CAB extends these results to deep feedforward networks. On the disjoint MNIST task CAB outperforms two other methods for coping with catastrophic interference that have recently been proposed in the deep learning field.

\newpage
\thispagestyle{empty}
\tableofcontents

\newpage
\setcounter{page}{1}

\section{Introduction}
When trained on a sequence of tasks individually, neural networks usually forget about previous tasks after the weights are adjusted for a new task. This notorious problem known as catastrophic interference \citep{french1999catastrophic, mccloskey1989catastrophic, kumaran2016learning, ratcliff1990connectionist} poses a serious challenge towards continual learning. An effective solution to this problem using conceptors was proposed by the second author \citep{Jaeger2014} in a reservoir computing context for incrementally loading dynamic patterns to a reservoir. Adopting and extending those methods, here we propose a \textit{conceptor-aided back-propagation} (CAB) algorithm to train feed-forward networks, and compare its performance to recent works that address the same problem.

The rest of this report is structured as follows. Section \ref{conceptor} introduces conceptors and their application to incremental learning by linear regression. Section \ref{CAB} extends the method to stochastic gradient descent and describes the CAB algorithm. Section \ref{experiment} demonstrates the performance of CAB on the permuted MNIST task and the disjoint MNIST task. Finally in Section \ref{discussion} we discuss the limitation of this method and its future extension.

\section{Incremental Ridge Regression by Conceptors}\label{conceptor}
In this section, we review the basics of conceptor theory and its application to incrementally training linear readouts of recurrent neural networks (RNNs) as used in reservoir computing. A comprehensive treatment can be found in \citep{Jaeger2014}. 

\subsection{Conceptors}
In brief, a \textit{matrix conceptor} $C$ for some vector-valued random variable $x\in \mathbb{R}^N$ is defined as a linear transformation that minimizes the following loss function.     
 \begin{align}
\label{matconobj}
\mathbb{E}_x[||x - Cx||^2]+\alpha^{-2}||C||_{\text{fro}}^2
\end{align}
where $\alpha$ is a control parameter called \textit{aperture} and $||\cdot||_{\text{fro}}$ is the Frobenius norm. This optimization problem has a closed-form solution 
\begin{align}
\label{conceptorsolution}
C = R(R+\alpha^{-2}I)^{-1}
\end{align}
where $R=\mathbb{E}_{x}[xx^{\top}]$ is the $N\times N$ correlation matrix of $x$, and $I$ is the $N\times N$ identity matrix. This result given in \eqref{conceptorsolution} can be understood by studying the singular value decomposition (SVD) of $C$. It has been shown that if $R=U\Sigma U^\top$ is the SVD of $R$,  then the SVD of $C$ is given as $U S U^\top$, where the singular values $s_i$ of $C$ can be written in terms of the singular values $\sigma_i$ of $R$: $s_i= \sigma_i/(\sigma_i+\alpha^{-2})\in [0,1)$. In intuitive terms, $C$ is a soft projection matrix on the linear subspace where the samples of $x$ lie, such that for a vector $y$ in this subspace, $C$ acts like the identity: $Cy\approx y$, and when some noise $\epsilon$ orthogonal to the subspace is added to $y$, $C$ de-noises: $C(y+\epsilon)\approx y$.  We define the quota $Q(C)$ of a conceptor to be the mean singular values: $Q(C):=\frac{1}{N}\sum_{i=1}^{N}s_i$. Intuitively, the quota measures the fraction of the total dimensions of the entire vector space that is claimed by $C$.

Moreover, logic operations that satisfy many laws of Boolean logic can be defined on matrix conceptors as the following:
\begin{align}
\label{bool}
\neg C:=& I-C,\\
C^i\vee C^j:=& (R^i+R^j)(R^i+R^j+\alpha^{-2}I)^{-1}\\
C^i\wedge  C^j:=&\neg(\neg C^i \vee \neg C^j) 
\end{align}

\subsection{Incremental Ridge Regression}
 With the help of these logic operations, conceptors can be applied to incrementally train one linear model by ridge regression for multiple input-to-output mapping tasks, such that (i) learning a new task does not interfere with previously learned tasks; (ii) similarities between tasks are exploited; (iii) the amount of remaining memory space can be monitored. Here ``memory space" refers to the linear space of input vectors. 

In particular, consider a sequence of $m$ incoming tasks indexed by $j$ and we denote the training dataset for the $j$-th task by $\{(x^j_1, y^j_1), \cdots, (x^j_n, y^j_n)\}$, where $x^j_i\in\mathbb{R}^N$ are input vectors and $y^j_i\in\mathbb{R}^M$ their corresponding target outputs. Whenever the training dataset for a new task is available, the incremental learning method will compute a matrix conceptor $C^j$ for the input variable of the new task and update the linear model, resulting in a sequence of linear models $W^1, \dots W^m$ such that $W^j$ solves not only the $j$-th task but also all previous tasks: for $k\leq j$, $y^k\approx W^jx^k$. The conceptor $C^j$ characterizes the memory space occupied by the $j$-th task, and we use $A^{j-1}=C^1\vee\cdots\vee C^{j-1}$ and $F^{j}=\neg A^{j-1}$ to represent the memory space already claimed by the tasks $1, \ldots, j-1$ and the memory space still free for the $j$-th task, respectively. More specifically, this method proceeds in the following way:

\begin{itemize}
	\item \textbf{Initialization (no task trained yet):} $W^{0}=0_{M\times N}, A^{0}=0_{N\times N}$. 
	\item \textbf{Incremental task learning:} For $j=1,\dots, m$ do:
	\begin{enumerate}
		\item Store the input vectors from the $j$-th training dataset of size $n$ into a $N\times n$ sized input collection matrix $X^j$, and store the output vectors into a $M\times n$ sized output collection matrix $Y^j$.
		\item Compute the conceptor for this task by $C^j = R^j(R^j+\alpha^{-2}I)^{-1}$, where $R^j=\frac{1}{n}X^{j}X^{j\top}$
		\item Train a matrix $W_{inc}^j$ (to be added to $W^{j-1}$, yielding $W^{j}$):
		\begin{enumerate}
			\item[(a)] $F^{j}:=\neg A^{j-1}$ (\textit{comment: this conceptor characterizes the ``still disposable" memory space for the $j$-th task}),
			\item[(b)] $T:=Y^j-(W^{j-1}X^j)$ (\textit{comment: this matrix consists of target values for a linear regression to compute $W_{inc}^j$}),
			\item[(c)] $S:=F^jX^j$ (\textit{comment: this matrix consists of arguments for the linear regression}),
			\item[(d)] $ W^j_{inc} = ((SS^\top/n+\lambda^{-2}I)^{-1} ST^\top/n)^\top$ (\textit{comment: carry out the regression, regularized by $\lambda^{-2}$}),
		\end{enumerate}
		\item Update $W^{j}$: $W^{j}=W^{j-1}+W^j_{inc}$.
		\item Update $A: A^j=A^{j-1}\vee C^j$ (\textit{comment: this is possible due to the associativity of the OR operation on conceptors})
	\end{enumerate}
\end{itemize}   

Intuitively speaking, when learning a new task, this algorithm leaves the already trained directions in the memory space (characterized by $A^{j-1}$) intact and exploits only the components of input vectors in the free space (characterized by $F^j$) to compensate errors for the new task. 

\section{Conceptor-Aided SGD and Back-prop}\label{CAB}
\subsection{SGD}
In the algorithm introduced in the previous section, $W_{inc}^j$ is computed by ridge regression, which gives the closed-form solution to minimize the cost function 
\begin{align}
\label{CostLMS}
\mathcal{J}(W_{inc}^j) := \mathbb{E}[|W_{inc}^js-t|^2]+\lambda^{-2}|W^j_{inc}|^2_\text{fro}
\end{align}
where $t = y^j-W^{j-1}x^j, s=F^jx^j$. One can also minimize this cost function by stochastic gradient descent (SGD), which starts from an initial guess of $W_{inc}^j$ and repeatedly performs the following update
\begin{align}
\label{SGD}
W_{inc}^j\leftarrow W_{inc}^j - \eta \nabla_{W_{inc}^j} \mathcal{J}(W_{inc}^j)
\end{align}
where $\eta$ is the learning rate and the gradient is given by:
\begin{align}
\label{gradientLMS}
\nabla_{W_{inc}^j} \mathcal{J}(W_{inc}^j) = 2\mathbb{E}[(W_{inc}^js-t)s^\top]+2\lambda^{-2}W_{inc}^j
\end{align}
 Substituting $t = y^j-W^{j-1}x^j, s=F^jx^j= (I-A^{j-1})x^j$ in \eqref{gradientLMS}, we get 
 \begin{equation} \label{substitute}
 \begin{aligned}[b]
 \nabla_{W_{inc}^j} \mathcal{J}(W_{inc}^j) &= 2\mathbb{E}[(W_{inc}^j(I-A^{j-1})x^j-y^j+W^{j-1}x^j)s^\top]+2\lambda^{-2}W_{inc}^j\\
  & = 2\mathbb{E}[(-W_{inc}^jA^{j-1}x^j+(W^{j-1}+W_{inc}^j)x^j-y^j)s^\top]+2\lambda^{-2}W_{inc}^j
 \end{aligned}
 \end{equation}
 Due to the regularization term in the cost function, as the optimization goes on, eventually $W_{inc}$ will null the input components that are not inside the linear subspace characterized by $F^{j}$, hence $W_{inc}^jA^{j-1}x^j$ will converge to 0 as the algorithm proceeds. In addition, since $W^j=W^{j-1}+W_{inc}^j$, \eqref{substitute} can be simplified to 
 \begin{align}
 \label{simplified}
 \nabla_{W_{inc}^j} \mathcal{J}(W_{inc}^j) = 2\mathbb{E}[(W^jx^j-y^j)s^\top]+2\lambda^{-2}W_{inc}^j
 \end{align}
 Adding $W^{j-1}$ to both sides of \eqref{SGD}, we obtain a update rule for $W^j$:
 \begin{align}
 \label{UpdateW}
 W^j\leftarrow W^j - 2\eta\mathbb{E}[es^\top]+2\eta\lambda^{-2}W_{inc}^j
 \end{align}
 where $e:=W^jx^j-y^j$. In practice, at every iteration, the expected value can be approximated by a mini-batch of size $n_B$, indexed by $i_B$:	
  \begin{align}
  \label{minibatch}
  \mathbb{\hat{E}}[es^\top] = \frac{1}{n_B}\sum_{i_B=0}^L (W^jx^j_{i_B}-y_{i_B}^j)(F^jx_{i_B}^j)^\top=\frac{1}{n_B}\sum_{i_B=0}^L (W^jx^j_{i_B}-y_{i_B}^j)x_{i_B}^{j\top}F^j
  \end{align}
  where the transpose for $F^j$ can be dropped since it is symmetric.
  
  If we only train the $j-$th task without considering the previous tasks, the update rule given by normal SGD will be 
  
   \begin{align}
   \label{UpdateW_LMS}
   W^j\leftarrow W^j - 2\eta\mathbb{E}[ex^{j\top}]+2\eta\alpha^{-2}W^j
   \end{align}
  Comparing this to the update rule in \eqref{UpdateW}, we notice two modifications when a conceptor is adopted to avoid forgetting: first, the conceptor-projected input vector $s=F^jx^j$ instead of the original input vector $x^j$ is used to calculate the gradient of weights; second, regularization is done on the weight increment $W_{inc}^j$ rather than the final weight $W^j$. These two modifications lead to our design of a conceptor-aided algorithm for training multilayer feed-forward networks.

\subsection{Backprop}
 The basic idea of CAB is to guide the gradients of the loss function on every linear component of the network by a matrix conceptor computed from previous tasks during error back-propagation, repeatedly applying the conceptor-aided SGD technique introduced in the previous section.

Consider a feed-forward network with $L+1$ layers, indexed by $l=0,\dots L$, such that the $0$-th and the $L$-th layers are the input and output layers respectively. $W^{(l)}$ represents the linear connections between the $(l-1)$-th and the $l$-th layer, where we refer to the former as the pre-synaptic layer with respect to $W^{(l)}$, and to the latter as the post-synaptic layer. We denote by $N^{(l)}$ the size of the $l$-th layer (excluding the bias unit) and ${A^{(l)}}^{j}$ a conceptor characterizing the memory space in the $l$-th layer used up by the first $j$ tasks. Let $\sigma(\cdot)$ be the activation function of the nonlinear neurons and $\theta$ all the parameters of the network to be trained. Then the incremental training method with CAB proceeds as follows:
\begin{itemize}
	\item \textbf{Initialization (no task trained yet):}
	 $\forall l= 0,\dots, L-1$, ${A^{(l)}}^0:=0_{(N^{(l)}+1)\times (N^{(l)}+1)}$, and randomly initialize ${W^{(l+1)}}^0$ to be a matrix of size $N^{(l+1)}\times (N^{(l)}+1) $.
	\item \textbf{Incremental task learning:} For $j=1,\dots, m$ do:
	\begin{enumerate}
		\item $\forall l= 0,\dots, L-1, {F^{(l)}}^j=\neg {A^{(l)}}^{(j-1)}.$ (\textit{This conceptor characterizes the still disposable memory space in layer $l$ for learning task $j$})
		\item Update the network parameters from $\theta^{(j-1)}$ to $\theta^{j}$ by stochastic gradient descent, where the gradients are computed by CAB instead of the classical backprop. Algorithms \ref{alg:forward} and \ref{alg:backward} detail the forward and backward pass of CAB, respectively. Different from classical backprop, the gradients are guided by a matrix conceptor ${F^{(l)}}^j$, such that in each layer only the activity in the still disposable memory space will contribute to the gradient. Note that the conceptors remain the same until convergence of the network for task $j$.
		\item After training on the $j$-th task, run the forward procedure again on a batch of $n_B$ input vectors, indexed by $i_B$, taken from the $j$-th training dataset, to collect activations ${h^{(l)}_{i_B}}^j$ of each layer into a $N^{(l)}\times n_B$ sized matrix ${H^{(l)}}^j$, and set the correlation matrix ${R^{(l)}}^j=\frac{1}{n_B}{H^{(l)}}^j({H^{(l)}}^j)^\top$.
		\item Compute a conceptor on the $l$-th layer for the $j$-th pattern by ${C^{(l)}}^j={R^{(l)}}^j({R^{(l)}}^j+\alpha^{-2}I_{N^{(l)}\times N^{(l)}})^{-1}, \forall l=0,\dots,L-1$. The aperture is set by trial and error, preferably in a cross-validation scheme.
		\item Update the conceptor for already used spaces in every layer: ${A^{(l)}}^j={A^{(l)}}^j\vee{C^{(l)}}^j, \forall l=0,\dots,L-1$.
	\end{enumerate}
\end{itemize}   
 
  \begin{algorithm}
  	\caption{The forward procedure of conceptor-aided backprop for the $j$-th task, adapted from \citep{Goodfellow-et-al-2016}. Input vectors are passed through a feed-forward network to compute the cost function. $\mathcal{L}(\hat{y}^j, y^j)$ denotes the loss for the $j$-th task, to which a regularizer $\Omega(\theta^j_{inc})=\Omega(\theta^j-\theta^{j-1})=||\theta^j-\theta^{j-1}||^2_{\text{fro}}$ is added to obtain the total cost $\mathcal{J}$, where $\theta$ contains all the weights (biases are considered as weights connected to the bias units). The update of parameters rather than the parameters themselves are regularized, similar to the conceptor-aided SGD.}\label{alg:forward}
  	\begin{algorithmic}[1]
  		\Require{Network depth, $l$}
  		\Require{${W^{(l)}}^j, l\in \{1,\dots, L\}$, the weight matrices of the network}
  		\Require{$x^j$, one input vector of the $j$-th task}
  		\Require{$y^j$, the target output for $x^j$ }
  		\State $h^{(0)}=x^j$
  		\For{$l=1,\dots L$} 
  		\State $b^{(l)}=[h^{(l-1)\top},1]^\top$, include the bias unit
  		\State $a^{(l)}={W^{(l)}}^jb^{(l)}$
  		\State $h^{(l)}=\sigma(a^{(l)})$
  		\EndFor
  		\State $\hat{y}^j=h^{(l)}$
  		\State $\mathcal{J}=\mathcal{L}(\hat{y}^j,y^j)+\lambda\Omega(\theta^j_{inc})$
  	\end{algorithmic}
  \end{algorithm}
  
  \begin{algorithm}
  	\caption{The backward procedure of conceptor-aided backprop for the $j$-th task, adapted from \citep{Goodfellow-et-al-2016}. The gradient $g$ of the loss function $\mathcal{L}$ on the activations $a^{(l)}$ represents the error for the linear transformation ${W^{(l)}}^j$ between the $(l-1)$-th and the $l-$th layers. In the standard backprop algorithm, the gradient of $\mathcal{L}$ on ${W^{(l)}}^j$ is computed as an outer product of the post-synaptic errors $g$ and the pre-synaptic activities $h^{(l-1)}$. This resembles the computation of the gradient in the linear SGD algorithm, which motivates us to apply conceptors in a similar fashion as in the conceptor-aided SGD. Specifically, we project the gradient $\nabla_{{W^{(l)}}^j}\mathcal{L}$ by the matrix conceptor ${F^{(l-1)}}^j$ that indicates the free memory space on the pre-synaptic layer.}\label{alg:backward}
  	\begin{algorithmic}[1]
  		\State
  		\begin{align*}
  		 g\leftarrow\nabla_{\hat{y}}\mathcal{J}=\nabla_{\hat{y}}\mathcal{L}(\hat{y}, y)
  		 \end{align*}
  		\For{$l=L,L-1,\dots, 1$}
  		\State Convert the gradient on the layer's output into a gradient on the pre-nonlinearity activation ($\odot$ denotes element-wise multiplication):
  		 \begin{align*}
  		 g\leftarrow\nabla_{a^{(l)}}\mathcal{J}=g\odot \sigma'(a^{(l)})
  		 \end{align*}
  		\State Compute gradients on weights, projected by ${F^{(l-1)}}^j$, and added to the regularization term on the increment:
  		 \begin{align*}
  		 \nabla_{{W^{(l)}}^j}\mathcal{J}=& g({F^{(l-1)}}^jb^{(l-1)})^\top+\lambda\nabla_{{W^{(l)}}^j}\Omega(\theta_{inc}^j)=g{b^{(l-1)}}^\top{F^{(l-1)}}^j+2\lambda{W^{(l)}_{inc}}^j\\
  		  = & g{b^{(l-1)}}^\top{F^{(l-1)}}^j+2\lambda({W^{(l)}}^j-{W^{(l)}}^{j-1})
  		 \end{align*}
  		\State Propagate the gradients w.r.t. the next lower-level hidden layer’s activations:
  		 \begin{align*}
  		 g\leftarrow \nabla_{h^{(l-1)}}\mathcal{J}={{W^{(l)}}^j}^\top g
  		  \end{align*}
  		\EndFor
  	\end{algorithmic}
  \end{algorithm}

\section{Experiments}\label{experiment}

\subsection{Disjoint MNIST Experiment}
 To test the performance of CAB, we applied it on the task of 10-class categorization for disjoint MNIST datasets \citep{srivastava2013compete, DBLP:journals/corr/LeeKHZ17}, where the original MNIST dataset is divided into two disjoint datasets with the first one consisting of data for the first five digits ($0$ to $4$), and the second one of the remaining five digits ($5$ to $9$). This task requires a network to learn these two datasets one after the other, then examines its performance of classifying the entire MNIST testing dataset into 10 classes. The current state-of-the-art accuracy on this task, averaged over 10 learning trials, is $94.12(\pm0.27)$, achieved by a method called Incremental Moment Matching (IMM) introduced in \citep{DBLP:journals/corr/LeeKHZ17}. They also reported the performance of Elastic Weight Consolidation (EWC) method proposed in \citep{kirkpatrick2017overcoming} to be $52.72(\pm 1.36)$ on this task.

For this task, we trained a feed-forward network with [784-800-10] neurons. Logistic sigmoid neurons are used in both hidden and output layers, and the network is trained with vanilla SGD to minimize mean squared error. An aperture $\alpha=9$ was used for all conceptors on all layers, learning rate $\eta$ and regularization coefficient $\lambda$ were chosen to be $0.1$ and $0.005$ respectively. The accuracy of CAB on this task, measured by repeating the experiment 10 times, is $94.91(\pm0.30)$.  It is important to mention that the networks used in \citep{DBLP:journals/corr/LeeKHZ17} for IMM and EWC had [784-800-800-10] neurons and rectified linear units (ReLU), so CAB achieved the state-of-the-art performance with less layers and neurons.

 \begin{figure*}
	\noindent\makebox[\textwidth]{%
		\begin{tabular}{c}
			\includegraphics[width=10cm]{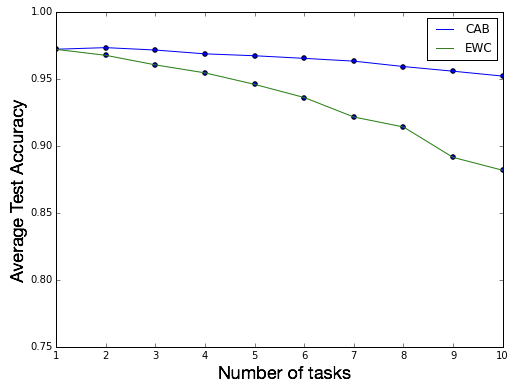} \\
		\end{tabular}
	}
	\caption{Average performance across different number of permuted MNIST tasks using CAB or EWC}
	\label{fig:permute10}
\end{figure*}

\subsection{Permuted MNIST Experiment}
 Another task on which we tested CAB is the permuted MNIST experiment \citep{goodfellow2013empirical, kirkpatrick2017overcoming, DBLP:journals/corr/LeeKHZ17, NIPS2013_5059}, where a sequence of pattern recognition tasks are created from the MNIST dataset \citep{lecun1998mnist}. For each task, a random permutation of input image pixels is generated and applied to all images in MNIST to obtain a new shuffled dataset, equally difficult to recognize as the original one, the objective of each task is to recognize these images with shuffled pixels.

For a proof-of-concept demonstration, we trained a simple but sufficient feed-forward network with [784-100-10] of neurons to classify 10 permuted MNIST datasets. Figure \ref{fig:permute10} shows the performance of CAB on this task, the average testing accuracy is 95.2 after learning all 10 tasks sequentially. The network has logistic sigmoid neurons in both hidden and output layers, and is trained with mean squared error as the cost function. Vanilla SGD was used in all experiments to optimize the cost function. Learning rate and aperture were set to 0.1 and 4, respectively. For comparison, we also tested EWC on the same task with the same network architecture, based on the implementation from \citep{EWC_code}. The parameters chosen for the EWC algorithm were 0.01 for the learning rate and 15 for the weight of the Fisher penalty term. Although a fair amount of effort was spent on optimizing the parameters of EWC, the accuracies shown here might still be far from its best performance.

 \begin{figure*}
 	\noindent\makebox[\textwidth]{%
 		\begin{tabular}{cc}
 			\includegraphics[width=12cm]{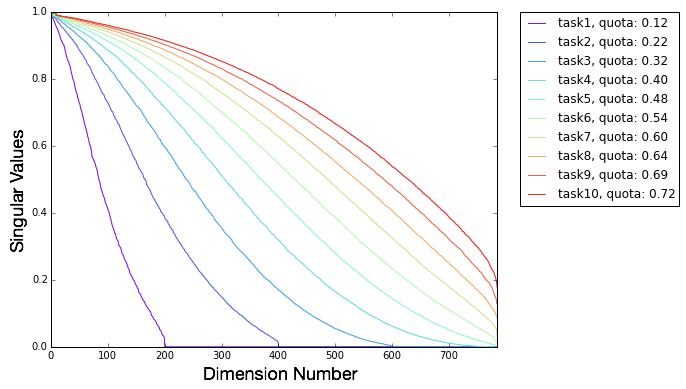} \\
 			(a) Singular value spectra of conceptors ${A^{(0)}}^j$ on the input layer.\\ \includegraphics[width=12cm]{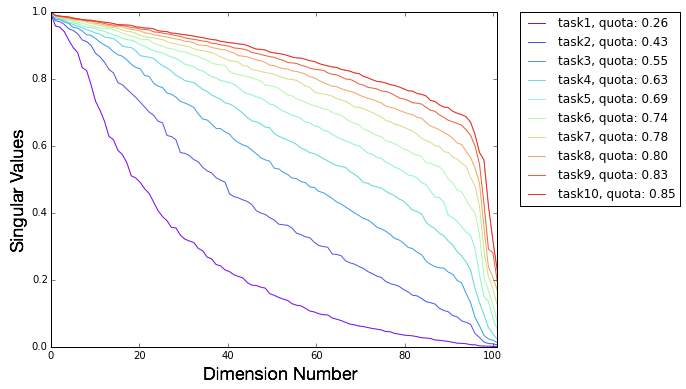}
 			\\ (b)  Singular value spectra of conceptors ${A^{(1)}}^j$ on the hidden layer.\\
 		\end{tabular}
 	}
 	\caption{The development of singular value spectra of conceptors for ``used-up" space on the input layer and hidden layer during incremental learning of 10 permuted MNIST tasks. Quota of these conceptors are displayed in the legends. }
 	\label{fig:spectra}
 \end{figure*}

Since all tasks are generated by permuting the same dataset, they should occupy the same portion of the input space. However, as more tasks are learned, the chance that the space of a new task will overlap with the already used input space increases. This can be seen clearly from Figure \ref{fig:spectra}, which shows the singular value spectra and the quota of conceptors for the already used spaces on the input and hidden layers respectively. As the incremental learning proceeds, it becomes less likely for a new task to be in the free space: the second task increases the quota of the input layer memory space by 0.1, whereas the 10th task increases it by only 0.03. However, CAB still manages to make the network learn new tasks based on their components in the non-overlapping space.

\section{Discussion}\label{discussion}

\begin{figure*}
	\noindent\makebox[\textwidth]{%
		\begin{tabular}{cc}
			(a) images projected by $C_{5to9}$  & (b) images projected by $F_{5to9}$ \\
			\includegraphics[width=4cm]{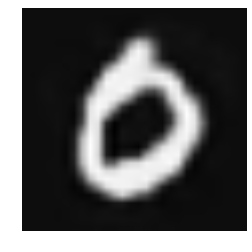} &
			\includegraphics[width=4cm]{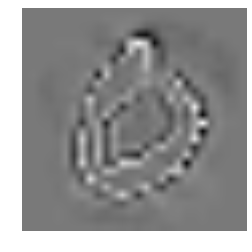}\\
			\includegraphics[width=4cm]{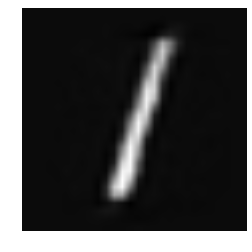} &
			\includegraphics[width=4cm]{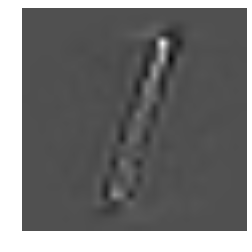}\\
			\includegraphics[width=4cm]{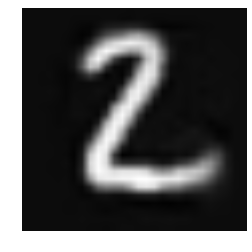} &
			\includegraphics[width=4cm]{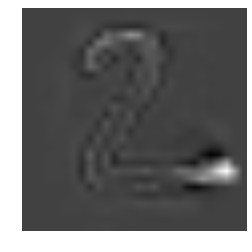}\\
			\includegraphics[width=4cm]{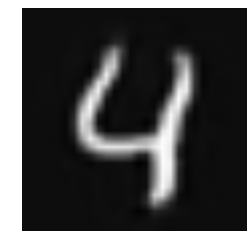} &
			\includegraphics[width=4cm]{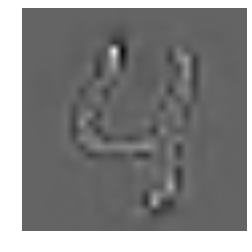}\\
		\end{tabular}
	}
	\caption{Projecting several images from the dataset for digits 0 to 4 by the conceptor $C_{5to9}$ and its negation $F_{5to9}$. After a network is trained on digits 5 to 9, CAB will only use the components projected by $F_{5to9}$ to correct the classification errors on these images.}
	\label{fig:disproj}
\end{figure*}

\begin{figure*}
	\noindent\makebox[\textwidth]{%
		\begin{tabular}{cc}
			(a) images projected by $C_{permuted}$  & (b) images projected by $F_{permuted}$ \\
			\includegraphics[width=4cm]{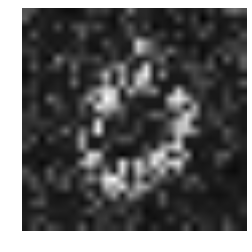} &
			\includegraphics[width=4cm]{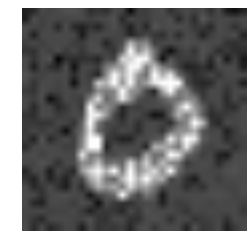}\\
			\includegraphics[width=4cm]{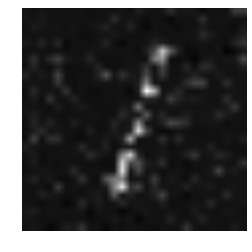} &
			\includegraphics[width=4cm]{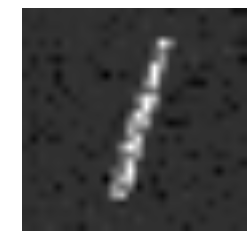}\\
			\includegraphics[width=4cm]{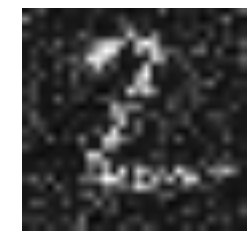} &
			\includegraphics[width=4cm]{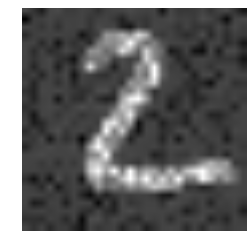}\\
			\includegraphics[width=4cm]{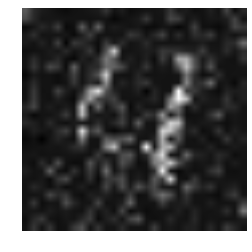} &
			\includegraphics[width=4cm]{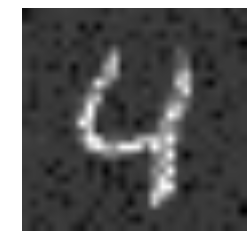}\\
		\end{tabular}
	}
	\caption{Projecting images from the original MNIST dataset by the conceptor $C_{permuted}$ computed from the shuffled MNIST dataset and its negation $F_{permuted}$. After a network is trained on the shuffled MNIST, CAB will only use the components projected by $F_{permuted}$ to correct the classification errors on these images.}
	\label{fig:perproj}
\end{figure*}

The experiment results from the previous section indicate that learning disjoint MNIST datasets is a more challenging task for CAB than learning the permuted MNIST datasets. To see why this is the case, it is important to understand that when learning a new task, CAB exploits only the components of input vectors that are not inside the linear subspace characterized by the conceptor of previous tasks. In other words, the more overlap there is between the conceptor of the new task and that of the already learned tasks, the less components in the input patterns are left for CAB to correct the network's errors on the new task, hence more difficult for the network to learn these tasks sequentially.

To quantify the overlap between two conceptors $C^i, C^j$, we can use the similarity measure proposed in \citep{Jaeger2014}:
\begin{align}
\label{Similarity}
\rho(C^i, C^j) = \frac{||(S^i)^{1/2}(U^i)^\top U^j(S^j)^{1/2}||_{\text{fro}}^2}{||\text{diag} S^i||\cdot||\text{diag} S^j||}
\end{align}
where $C^i = U^iS^i (U^i)^\top$ and $C^j = U^jS^j (U^j)^\top$ are their singular value decompositions. This measure ranges in $[0,1]$. It is 0 if and only if $C^i, C^j$ specify two orthogonal linear subspaces, and 1 if and only if $C^i$ is a multiple of $C^j$.

In order to compare the difficulties of the permuted and disjoints MNIST experiments, we selected four datasets: $D_{original}, D_{permuted}, D_{5to9}$ and $D_{0to4}$, where $D_{original}$ is the original MNIST dataset; $D_{permuted}$ is the whole MNIST dataset but the pixels of every image is shuffled by the same randomly generated permutation; $D_{5to9}$ consists of only the images of digits $5$ to $9$ from the MNIST dataset, and $D_{0to4}$ has only the images of digits $0$ to $4$. Then for each of these four datasets, we computed a conceptor from the raw input images data inside it. The results were four conceptors $C_{original}, C_{permuted}, C_{5to9}$ and $C_{0to4}$.

In the permuted MNIST experiment, the network has to learn to recognize $D_{permuted}$ and $D_{original}$ sequentially, the overlap between their corresponding conceptors can be measured by $\rho(C_{original}, C_{permuted})$, which is around 0.3 on average.

In contrast, $\rho(C_{5to9}, C_{0to4})$ is much higher ($\approx0.95$), which means the input images in $D_{5to9}$ and $D_{0to4}$ span roughly the same linear subspaces of the input memory space. Therefore, if a network is first trained on $D_{5to9}$ and then on $D_{0to4}$, only a very small amount of components of the images in $D_{0to4}$ can be exploited to compensate the errors, namely those components preserved by $F_{5to9} := \neg C_{5to9}$, whereas the linear transformation of the components inside the subspace characterized by $C_{5to9}$ will be fixed. Figure \ref{fig:disproj} shows some images from $D_{0to4}$ projected by $C_{5to9}$ and $F_{5to9}$. Note that when learning to recognize the second dataset, CAB only allows the network to adjust its performance based on the projected versions displayed in the right column, which are much less legible than the images projected by $C_{5to9}$, shown in the left column. 

Since the images in Figure \ref{fig:disproj} are also included in $D_{original}$, for comparison, we also visualized their components inside the linear subspaces characterized by $C_{permuted}$ and $F_{permuted}$, which can be found in Figure \ref{fig:perproj}. In the setting of permuted MNIST experiment, after the network is trained on $D_{permuted}$, it can only rely on the components displayed in the right column of Figure \ref{fig:perproj} to compensate its output errors. However, it is clear that the images in the right column of Figure \ref{fig:perproj} are more distinguishable than those in the right column of Figure \ref{fig:disproj}, hence the permuted MNIST experiment is easier for CAB than the disjoint one.

A direction for future work, suggested by the analysis above, is to improve
CAB such that the weights of the network can be adjusted even when the input patterns of different tasks lie in the same linear subspace. On the other hand, such similarity between tasks might be exploited to save the training time. So another question for further investigation is how to turn the similarity between different tasks into a desirable property rather than difficulty for continual learning.\\*

{\bf Acknowledgments.} The work reported in this article was partly funded through the European H2020 collaborative project NeuRAM3 (grant Nr 687299).

\bibliographystyle{apalike}
\bibliography{forget}

\noindent

\end{document}